\pdfoutput=1

\documentclass[11pt]{article}

\usepackage[final]{acl}

\usepackage{times}
\usepackage{latexsym}
\usepackage{hyperref}
\usepackage{url}
\usepackage{graphicx}
\usepackage{booktabs}
\usepackage{algorithm}
\usepackage{algorithmic}
\usepackage{amsmath}
\usepackage{amssymb}
\usepackage{array}
\usepackage{lipsum}
\usepackage{multirow}
\usepackage{soul}
\usepackage{wrapfig}
\usepackage{enumitem}
\usepackage{bbding}

\usepackage[T1]{fontenc}

\usepackage[utf8]{inputenc}

\usepackage{microtype}

\usepackage{inconsolata}

\usepackage{graphicx}

\title{Alignment-Augmented Speculative Decoding with Alignment Sampling and Conditional Verification}

\author{
 \textbf{Jikai Wang\textsuperscript{1,}\textsuperscript{2}},
 \textbf{Zhenxu Tian\textsuperscript{1,}\textsuperscript{2}},
 \textbf{Juntao Li\textsuperscript{1,}\textsuperscript{2}\thanks{Corresponding author.}},
\\
 \textbf{Qingrong Xia\textsuperscript{3}},
 \textbf{Xinyu Duan\textsuperscript{3}},
 \textbf{Zhefeng Wang\textsuperscript{3}},
 \textbf{Baoxing Huai\textsuperscript{3}},
 \textbf{Min Zhang\textsuperscript{1,}\textsuperscript{2}},
\\
 \textsuperscript{1}Soochow University, China\\
 \textsuperscript{2}Key Laboratory of Data Intelligence and Advanced Computing, Soochow University\\
 \textsuperscript{3}Huawei Cloud,China
\\
    \href{mailto:risus254@gmail.com}{risus254@gmail.com},
    \href{mailto:ljt@suda.edu.cn}{ljt@suda.edu.cn}
 }

\begin{document}
\maketitle
\begin{abstract}
Recent works have revealed the great potential of speculative decoding in accelerating the autoregressive generation process of large language models.
The success of these methods relies on the alignment between draft candidates and the sampled outputs of the target model.
Existing methods mainly achieve draft-target alignment with training-based methods, e.g., EAGLE, Medusa, involving considerable training costs.
In this paper, we present a training-free alignment-augmented speculative decoding algorithm. 
We propose alignment sampling, which leverages output distribution obtained in the prefilling phase to provide more aligned draft candidates.
To further benefit from high-quality but non-aligned draft candidates, we also introduce a simple yet effective flexible verification strategy. 
Through an adaptive probability threshold, our approach can improve generation accuracy while further improving inference efficiency.
Experiments on 8 datasets (including question answering, summarization and code completion tasks) show that our approach increases the average generation score by 3.3 points for the LLaMA3 model.
Our method achieves a mean acceptance length up to 2.39 and speed up generation by 2.23×.





\end{abstract}

\section{Introduction}
The enormous size of state-of-the-art autoregressive models ~\citep{cluade,llama31,openai2024reason} demands substantial memory and processing power, making real-time applications challenging. 
In scenarios such as interactive text generation, these models require vast amounts of computation, leading to slower response times and increased energy consumption.
Hence, more efficient decoding algorithms are urgently needed to reduce inference costs significantly while maintaining or improving generation quality. 

Speculative decoding (SD) \citep{stern2018blockwise,leviathan2023fast,xia2023speculative,xia-etal-2024-unlocking,li2024eagle} has emerged as an effective solution to the inefficient decoding process of large autoregressive models. 
It uses a ``draft and verify'' mechanism to achieve lossless acceleration. 
A lightweight drafting model generates candidate tokens, which are then verified in parallel by the target model. 
This enables the generation of multiple tokens in a single step, thus accelerating decoding while preserving the output distribution. 
Speculative decoding algorithms that combine large and small models \citep{miao2023specinfer, li2024eagle} have shown promising results. 
However, these approaches often require additional training and extra parameters, complicating deployment in inference systems \citep{2023lmdeploy, kwon2023efficient}.

As an alternative, the training-free retrieval-based speculative decoding \citep{yang2023inference, saxena2023prompt, he-etal-2024-rest} uses external knowledge sources such as databases or historical text to retrieve n-grams as drafts for generation. 
It is more flexible and scalable to large models, as the drafting cost is independent of model size. 
However, two issues have been largely overlooked in previous work on retrieval-based speculative decoding. 
First, poor alignment between the retrieved drafts and the model’s output distribution leads to a low acceptance rate. 
Second, in many generation scenarios, the input context and the generated content have strong correlation.
As a result, drafts retrieved from the input context tend to be of higher quality and do not require strict verification to ensure consistency with the target model's output.

In this paper, we propose \textbf{A}lignment-\textbf{A}ugmented \textbf{S}peculative \textbf{D}ecoding (AASD), a plug-and-play generation algorithm that improves the draft-target alignment in prompt-based speculative decoding. 
To address the first issue, we introduce alignment sampling, where we sample additional tokens from the output distribution of the prefilling phase for poorly aligned tokens, thereby improving the overall draft quality and increasing the chances of successful alignment with output distribution of the target model.
To address the second issue, we propose a conditional verification strategy. 
It applies heuristic probability thresholds for each token, based on its information entropy, allowing the model to autonomously utilize the input text during decoding through speculative decoding. 
It improves both generation accuracy and efficiency.
Both the two techniques improve the alignment between the retrieved draft candidates and the target model output distribution. 
Alignment sampling makes the drafts more aligned with the target model distribution, and conditional verification makes the target model more aligned with the high-quality drafts.

We conduct comprehensive experiments with two different models on 8 datasets in long context generation scenarios, including question answering, summarization, and code completion tasks, in Section \ref{sec:exp} to evaluate the effectiveness of AASD. 
The results show that AASD outperforms common sampling methods in terms of generation performance, improving the average score from 44.69 to 47.98 compared to greedy sampling for the LLaMA3 model \citep{llama31}.
In terms of generation efficiency, AASD achieves the highest acceptance rate and average decoding throughput among existing retrieval-based speculative decoding algorithms. 
It achieves a decoding speedup of up to 2.23 times.

In summary, we highlight the importance of draft-target alignment in retrieval-based speculative decoding in this paper. 
We propose a training-free method called alignment-augmented speculative decoding to improve this alignment, which consists of alignment sampling and conditional verification. 
The proposed method improves generation accuracy and efficiency compared to baseline methods.


\section{Backgrounds}
\subsection{Speculative Decoding}
Given a prompt $q=(x_{1},x_{2},...,x_{l})$ and language model $M$, where $x_{i(i=1,2,...,l)}$ represents each token, and $l$ is the prefix length, we input the current sequence into the model to obtain the next token in the autoregressive decoding manner:
\begin{equation}
y_{l+1}\sim p_{M}(x_{l+1}|(x_{1},x_{2},...,x_{l})),
\end{equation}
where $p_{M}( \cdot |\cdot)$ represents the output probability distribution of $M$.
The model can generate only one token in one forward propagation, resulting in low efficiency, particularly for long generation lengths..

Speculative decoding (SD) alleviates this problem by adopting a ``draft and then verify'' mechanism. 
At each decoding step, a lightweight model $m$ generates a draft $d=(\hat{x}_{l+1},\hat{x}_{l+2},...,\hat{x}_{l+l_{d}})$ for the next tokens based on the current input sequence, where $\hat{x}_{i}(i=l+1,l+2,...,l+l_{d})$ denotes the draft tokens and $l_{d}$ represents the draft length.
This draft is then simultaneously input into the target model for verification.
Tokens in the draft that match the target model's output will be accepted.
Note that if a token is rejected, all subsequent tokens in the draft are also rejected.
This allows the model to generate multiple tokens in a single forward pass.

\subsection{Retrieval-Based Drafting}
\label{sec:pre}
An effective draft model must address two crucial aspects.
First, it must minimize computational costs, ensuring faster and more efficient inference through a lightweight design. 
Second, it must closely align with the target model.
Ideally, perfect alignment would result in a 100\% acceptance rate, thus achieving the theoretical maximum speedup, defined as the throughput of the smaller model divided by that of the target model.

Approaches such as EAGLE \citep{li2024eagle} strengthen the alignment between the draft and target models through alignment training, causing inconvenience in adapting to different models.
Retrieval-based speculative decoding methods complete the drafting by retrieving n-grams from historically generated data or pre-provided databases based on the current sequence.
However, existing retrieval-based speculative decoding methods \citep{yang2023inference,saxena2023prompt, he-etal-2024-rest} ignore the importance of alignment. 
Although their drafting overhead is very small compared to methods that require additional modules, their acceptance rate is also relatively lower.

\subsection{Impact of Draft-Target Alignment}
\begin{figure}[t]
    \centering
    \includegraphics[width=\columnwidth]{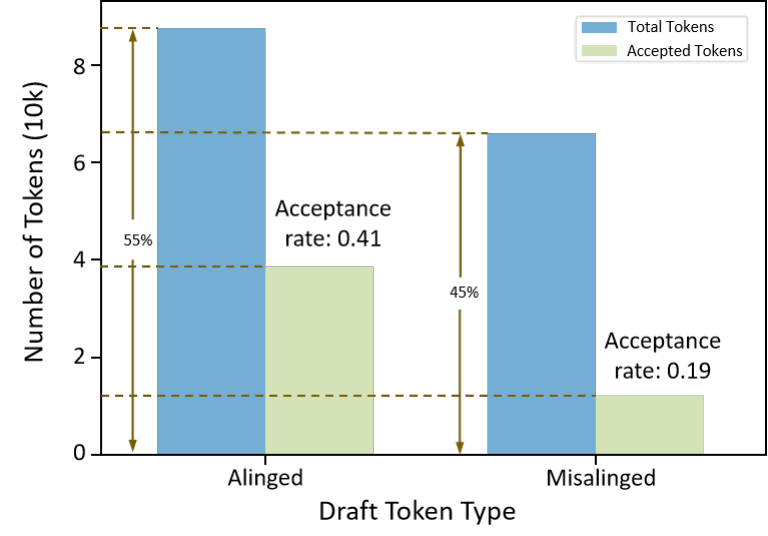}
    \caption{\label{fig:alignment} Acceptance rate of aligned and misaligned draft tokens in prompt-based speculative decoding.}
\end{figure}
Recent research \citep{anonymous2024judge} shows that even a high-quality draft will have a low acceptance rate if it does not align with the model output distribution.
We also conduct an experiment to give an insight into the impact of draft-target alignment in retrieval-based speculative decoding.
For the four-token input sequence ``A, B, C, D'', the model can predict the probability distribution of the next token for each input token in parallel. For each distribution, we select the token with the highest probability, that is, greedy sampling. Assume that the sampled tokens are B, C, F, E in order. The sampled tokens ``B'' and ``C'' match the input, while ``F'' does not match ``D'' in the input. We regard the matching tokens (``B'' and ``C'') as aligned tokens and the unmatched tokens (``D'') as misaligned tokens.
We conduct prompt-based speculative decoding on 100 samples in NQ \citep{47761} dataset with LLaMA3.1-8B-Instruct \citep{llama31} and calculate the acceptance frequency of aligned and misaligned draft tokens.
Results are shown in Figure \ref{fig:alignment}.
About 45\% retrieved draft tokens are misaligned with the model output.
The acceptance rate of misaligned tokens is much lower than that of aligned tokens.
Inspired by this, for the positions of misaligned tokens, we can sample additional tokens from the output distribution to enhance the alignment of the entire draft and the model, thus improving the acceptance rate.

\subsection{Potential for Non-Strict Verification}
Most speculative decoding methods adopts strict verification to achieve lossless generation acceleration.
The recent research \citep{anonymous2024judge} claims that correct but non-aligned draft candidates can be accepted to pursue higher speed-up ratio.
They trained an additional module on carefully designed data to judge whether to accept draft tokens.
However, for prompt-based speculative decoding, as the draft source is highly relevant to the answer, a training-free verification strategy can be designed to realize a better trade-off between generation performance and efficiency.
For example, in long context generation scenario, the input context contains the key information needed for model generation.
Some of the fragments even overlap with the standard generated content.
Given this, we can design a conditional verification method to allow the model to accept high-quality draft tokens, rather than just accept that match its original output.
This allows the model to adaptively call the input original segments on the decoding side, thus enhancing the generation accuracy and efficiency.

\begin{figure*}[t]
    \centering
    \includegraphics[width=\textwidth]{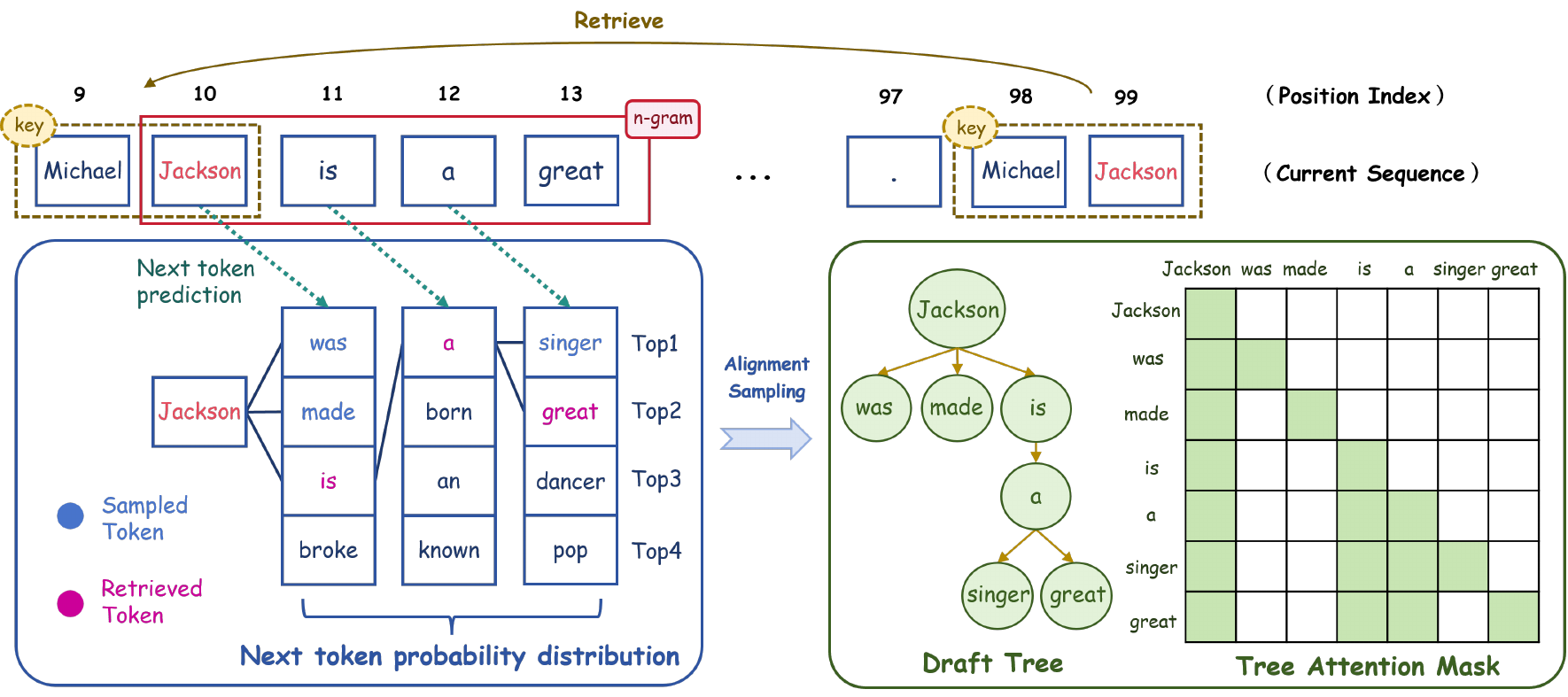}
    \caption{\label{fig:as} Illustration of alignment sampling. ``Jackson is a great'' is an n-gram retreived by the key ``Michael Jackson''. According to the output distribution obtained in the prefilling stage, ``is'' is the token with the third highest probability. The top-1 token ``was'' and top-2 token ``made'' are more likely to be generated after ``Michael Jackson''. Therefore, we sample two more tokens (``was'' and ``made'' for the 100th position). Token ``a'' is well aligned with the model output, so we keep it unchanged for the 101st position.}
\end{figure*}

\section{Method}
\label{sec:method}
In this section, we introduce AASD, an alignment-augmented speculative decoding algorithm based on alignment sampling and conditional verification that improves both generation performance and efficiency.
It is a plug-and-play speculative decoding algorithm that does not require additional modules or training.
The overall description of AASD is dispalyed in Algorithm \ref{alg:AASD} in Appendix \ref{sec:alg}.

\subsection{Context-Aware Drafting}
For long text generation, the input context can act as an important source of draft, because it is highly relevant to the generated content.
We construct the input of each request into a draft pool by sliding window sampling, which is an index dictionary.
The keys in the draft pool are several consecutive tokens in the input context and the values are the position indexes where these tokens appear.
We update the draft pool when new tokens are generated.

In each decoding step, we use the last several tokens of the current sequence as the key to retrieve indexes in the draft pool.
Since longer keys give higher draft relevance, following REST \citep{he-etal-2024-rest}, we prioritize searching with longer keys and obtain n-grams from the current sequence.

\subsection{Alignment Sampling}
As is shown in Figure \ref{fig:as}, we perform alignment sampling to expand the retrieved n-grams into draft trees.
For example, the token ``is'' is the third most likely word in the probability distribution of the next word of the token ``Jackson''.
Therefore, ``was'' and ``made'' are potential high-quality draft tokens, which are more aligned with the output distribution of the target model.
We additionally sample these two tokens for the 100th position.
In this way, we expand and merge multiple candidate n-grams into a draft tree and calculate the corresponding tree attention mask for parallel verification.
The draft tree $\mathbb{T}$ is defined as:
\begin{equation}
 \mathbb{T}=( \mathbb{V},\mathbb{E}),\mathbb{V}=\bigcup_{i=l+1}^{l+r}\bigcup_{j=1}^{n_{i}}\left\{ \hat{x}^{j}_{i}\right\},
\end{equation}
where $\mathbb{V}$ and $\mathbb{E}$ is the set of its nodes and edges.
$n_{i}$ is the number of retrieved tokens in the $i_{th}$ layer of $\mathbb{T}$.
$r$ is the depth of $\mathbb{T}$.
Then we input $\mathbb{T}$ and the corresponding tree attention mask into the model and get the output probability distribution $P_{M}(x_{i}|\mathbb{F}(\hat{x}^{j}_{i}))$ for each node, where $\mathbb{F}(\hat{x}^{j}_{i})$ is the set of all parent nodes (including prefix) of $\hat{x}^{j}_{i}$.

\subsection{Conditional Verification}
As discussed in Section \ref{sec:pre}, high-quality drafts retrieved from prompts have the potential to be non-rigorously verified to enhance generation accuracy.
The naive idea is to use a fixed probability threshold to filter draft tokens or use top-\textit{k} verification.
We argue that using the probability threshold condition is superior to using the top-\textit{k} condition alone since the output probability directly reflects the model’s confidence in each draft token.
Moreover, the verification condition of top-\textit{k} is not applicable for some extreme cases.
For example, suppose the probability of a certain token is close to 1, and the probability of all other tokens is close to 0. In that case, no other token should be accepted, except for the token with the highest probability.
If all tokens have the same probability, then they should all be accepted or rejected at the same time.
However, the top-\textit{k} verification condition will only pass individual tokens and reject others.
Therefore, we first set a probability threshold $\delta$ so that token $\hat{x}^{j}_{i}$ will be accepted if the following conditions are met:
\begin{equation}
p_{M}(\hat{x}^{j}_{i}|\mathbb{F}(\hat{x}^{j}_{i}))\geqslant \delta.
\end{equation}

However, models exhibit varying levels of confidence across different samples or tokens. A static threshold may either over-restrict or under-restrict the verification process, leading to suboptimal performance. 
Therefore, we formulate an adaptive threshold based on the verification probability distribution.
Low-confidence tokens can be processed with stricter thresholds to maintain precision, while high-confidence tokens benefit from relaxed thresholds to ensure recall.
Therefore, we adjust the verification threshold of each token based on the information entropy of its verification probability distribution.
Draft token $\hat{x}^{j}_{i}$ is accepted under the following condition:
\begin{equation}
p_{M}(\hat{x}^{j}_{i}|\mathbb{F}(\hat{x}^{j}_{i}))\geqslant \delta^{j}_{i},
\end{equation}
where $\delta^{j}_{i}$ is calculated by:
\begin{equation}
\delta^{j}_{i}= min(-\alpha \sum P(\hat{x}^{j}_{i}) \log P(\hat{x}^{j}_{i})+\beta, \Delta ),
\end{equation}
where
\begin{equation}
\Delta=\max_{\hat{x}^{j}_{i}} P(\hat{x}^{j}_{i}),
\end{equation}
$\alpha$ and $\beta$ are hyperparameters.
$\alpha$ a is the factor affecting the threshold value due to information entropy.
$\beta$ is the benchmark value of the threshold.
The function of $\Delta$ is to ensure that the token with the highest probability will be accepted during verification.
Considering that the overall quality of the alignment sampled tokens is lower than the tokens retrieved from the prompt.
We only perform conditional verification on the draft token retrieved from the input context and selected by alignment sampling.
Strict verification is still applied for tokens retrieved from generated context.
We accept the longest draft candidate that passes verification.

\begin{table*}[ht!]
\centering
\small
\renewcommand{\arraystretch}{1.1}
\resizebox{\textwidth}{!}{
\begin{tabular}{lccccccccc}
\toprule
\multirow{2}{*}{\textbf{Method}}      & \textbf{NQ} & \textbf{TQA} & \textbf{2WikiMQA} & \textbf{HotpotQA} &  \textbf{Multi-News} &\textbf{GovReport} & \textbf{RepoBench-P} & \textbf{LCC} & \multirow{2}{*}{\textbf{Avg.}} \\
\cmidrule{2-9}
 & F1 & F1 & F1 & F1 &  ROUGE-L & ROUGE-L & Edit Sim & Edit Sim & \\
\midrule
\multicolumn{10}{c}{LLaMA3.1-8B-Instruct} \\
\midrule
\textbf{Greedy}      &  29.13 &    \textbf{92.13} &     42.83 &     49.80 &       13.51      &     15.96        &  48.11   &  66.05 & 44.69  \\
\textbf{Top-\textit{k}}  &  29.08 &  84.52   &    38.48 &     42.01 &       13.34      &     15.18        &  45.36   & 56.19 & 40.52  \\
\textbf{Nucleus}      &  29.21 &    86.39 &    39.98 &     44.79 &       13.58     &     15.46        &  45.59   &  55.82 & 41.35  \\
\textbf{Beam}      &  32.57 &   83.04 &   43.29 &     48.93 &       14.07      &     17.75        &  46.27   &  67.52 & 44.16  \\
\textbf{AASD}       & \textbf{34.54}  &  91.56   &   \textbf{43.10}   &   \textbf{50.38}   &    \textbf{15.05}      &      \textbf{19.34}       &  \textbf{58.84} & \textbf{71.03} & \textbf{47.98} \\
\midrule
\multicolumn{10}{c}{Qwen2.5-32B-Instruct} \\
\midrule
\textbf{Greedy}    & 41.90 &  81.90  &  32.35   & 35.73  &  \textbf{12.77}    &        14.58    &   37.72  & 60.85 & 39.72 \\
\textbf{Top-\textit{k}}    & 40.86 &  80.02  &  28.17   & 36.09  &  11.72       &      13.19     &  36.14   & 50.56 & 37.09 \\
\textbf{Nucleus}     & 41.64 & 79.07   &  29.64   & 35.37  &    12.30     &      13.56      &  38.01   & 49.48 & 37.43 \\
\textbf{Beam}     & 36.78 &  \textbf{84.57}  &  17.49   & 19.55  &  9.49       &    11.07        &   \textbf{42.04}  & \textbf{63.26} & 35.53 \\
\textbf{AASD}     & \textbf{43.34} &  83.55  &  \textbf{34.99}   & \textbf{39.12}  &  12.67       &        \textbf{15.25}    &    37.23  & 58.65 & \textbf{40.60} \\
\bottomrule
\end{tabular}
}
\caption{\label{tab:main} Performance of different sampling methods with LLaMA3.1-8B-Instruct and Qwen2.5-32B-Instruct. For all metrics, higher scores indicate better performance. The best results are in bold.}
\end{table*}

\begin{table*}[h!]
\centering
\small
\renewcommand{\arraystretch}{1}
\begin{tabular}{lccccccccc} 
\toprule
\multirow{2}{*}{\textbf{Method}}                   & \multicolumn{3}{c}{\textbf{2WikiMQA}} & \multicolumn{3}{c}{\textbf{GovReport}} & \multicolumn{3}{c}{\textbf{LCC}}  \\ 
\cmidrule{2-10}
     & MAL & TPS  & Speed-up           & MAL & TPS  & Speed-up                   & MAL & TPS  & Speed-up              \\
\midrule
\multicolumn{10}{c}{LLaMA3.1-8B-Instruct} \\
\midrule
\textbf{ARD}         &  1.00  &   20.10    &  1.00            &  1.00   &  20.15  &    1.00                   &   1.00  &    20.67 &         1.00        \\
\textbf{REST}      &   1.08  &   18.26   &  0.91    &  1.27  & 21.20  & 1.05 &    1.26   &   26.09  &  1.07    \\
\textbf{PLD}  &  1.93  &  32.77   &  1.63   &  1.42  & 21.33  & 1.06 & 2.06      &   36.52  &  1.77    \\
\textbf{AASD}     &  \textbf{2.37}   &  \textbf{37.18}  &  \textbf{1.85}     &  \textbf{1.97}  &  \textbf{34.00} & \textbf{1.69} &      \textbf{2.39}     & \textbf{46.20}   &    \textbf{2.23}  \\
\midrule
\multicolumn{10}{c}{Qwen2.5-32B-Instruct} \\
\midrule
\textbf{ARD}         &  1.00  &   8.08   &  1.00            &  1.00   &  8.14  &    1.00                   &   1.00  &   7.35  &         1.00        \\
\textbf{REST}      &  1.07   &  7.33  &  0.91   &  1.26 & 8.16 & 1.00 &  1.16   &  7.53 &  1.02   \\
\textbf{PLD}    &  2.13   &  \textbf{14.04}  &  \textbf{1.74}   &  1.51 & 10.01 & 1.23 &  1.46   & 8.72  &  1.19   \\
\textbf{AASD}    &  \textbf{2.14}   &  13.87  &  1.72   & \textbf{2.18}  & \textbf{14.20} & \textbf{1.74} &  \textbf{1.89}   &  \textbf{11.80} &  \textbf{1.61}   \\
\bottomrule
\end{tabular}
\caption{\label{tab:speed} Inference efficiency of AASD with LLaMA3.1-8B-Instruct and Qwen2.5-32B-Instruct. ARD refers to auto-regressive decoding with greedy sampling. MAL represents mean acceptance length and TPS represents tokens per second. The best results are in bold.}
\end{table*}

\section{Experiments}
\label{sec:exp}
\subsection{Settings}
In most speculative decoding methods with strict verification (e.g., REST \citep{he-etal-2024-rest}, PLD \citep{saxena2023prompt} and LLMA \citep{yang2023inference}), the performance on datasets depends solely on the sampling strategy rather than the specific decoding algorithm.
Equipped with conditional verification, AASD introduces a novel sampling method, which enhances access to input fragments.
Therefore, we compare the performance of \textbf{AASD} with several common sampling methods: 1) \textbf{Greedy Sampling}: It samples the token with the highest probability; 2) \textbf{Top-\textit{k} Sampling} \citep{fan-etal-2018-hierarchical}: It selecting the next token from the top \textit{k} most probable options ($k=50$); \textbf{Nucleus Sampling} \citep{Holtzman2020The}: It is also called top-\textit{p} sampling ($p=0.8$), which selects the next token from the smallest set of top tokens whose cumulative probability exceeds a predefined threshold; 3) \textbf{Beam Search} \citep{li-etal-2016-deep}: It explores multiple candidate sequences at each step, keeping only the top \textit{n} most likely options (the beam width $n=10$), to find an optimal or near-optimal output sequence.

On the other hand, we compare the performance and efficiency of \textbf{AASD} with autoregressive decoding and other retrieval-based speculative decoding methods.
Baselines are as follows: 1) \textbf{Auto-regressive decoding (ARD)}, 2) \textbf{REST} \citep{he-etal-2024-rest}: It retrieves n-grams from an external general database and constructs a tree-structured draft, 3) \textbf{PLD} \citep{saxena2023prompt}: It retrieves n-grams directly from the input context.
Note that \textbf{LLMA} \citep{yang2023inference} and \textbf{PLD} can be considered as the same method since they are very similar.

We evaluate the proposed method with LLaMA3.1-8B-Instruct \citep{llama31} and Qwen2.5-32B-Instruct \citep{qwen2025qwen25technicalreport} on 3 different tasks: 1) \textbf{Question Answering (QA)}: Nature Question (NQ) \citep{47761}, TriviaQA (TQA) \citep{joshi-etal-2017-triviaqa}, 2WikiMQA \citep{ho2020constructing} and HotpotQA \citep{yang2018hotpotqa}; 2) \textbf{Summarization}: Multi-News \citep{fabbri-etal-2019-multi} and GovReport \citep{huang2021efficient}; 3) \textbf{Code Completion}: RepoBench-P \citep{liu2024repobench} and LCC \citep{guo2023longcoder}.
For the NQ dataset, we randomly sampled 300 queries from the validation set for evaluation. We use subsets sampled by LongBench \citep{bai2023longbench} for other datasets.
We report F1 score for QA datasets, ROUGE-L for summarization datasets and Edit Sim for code completion datasets.
We also test the decoding efficiency on \textbf{SpecBench} \citep{xia-etal-2024-unlocking}.

We use the same generation hyperparameters for all baselines.
REST involves an additional database as the draft source. We selected the best database from the officially released database for testing (for each model and each datasets).
We did not compare with other training-required methods for generating enhancements and speculative algorithms because they require additional training or use additional models.
We use 6-grams and set the maximum length of the key for retrieval to 6 for AASD.
We use unified hyperparameters for AASD on all datasets.
$\alpha$ is set to 0.1 and 0.2 for LLaMA3.1 and Qwen2.5, respectively.
$\beta$ is set to 0.1.
Each position can be extended by at most two tokens by alignment sampling during drafting.
We use 4 A100-PCIE-40GB GPUs for all experiments.

\begin{figure}[t!]
    \centering
    \includegraphics[width=\columnwidth]{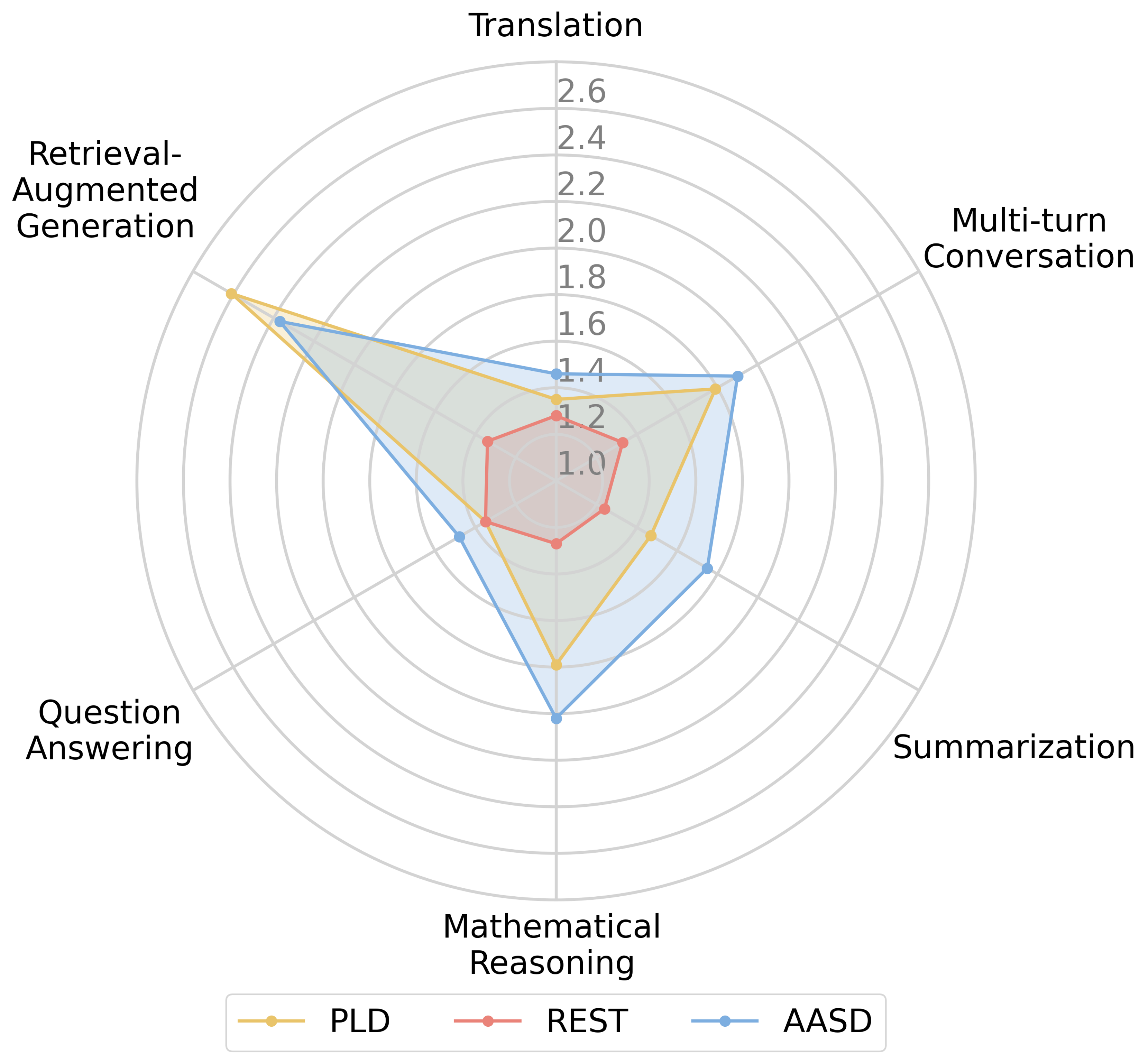}
    \caption{\label{fig:radar} Speed-up ratio on the six different task categories in SpecBench. The center point (1.0×) represents the baseline of autoregressive decoding.}
    \vspace{-10pt}
\end{figure}

\subsection{Improvement on Accuracy}
\label{sec:acc}

We compare the performance of different sampling methods in Table \ref{tab:main}.
We report the average scores of the 8 datasets in the last column.
AASD outperforms other sampling methods with LLaMA3.1-8B-Instruct on these datasets except for TQA.
It improves the average score from 44.69 to 47.98 compared to greedy sampling.
For Qwen2.5-32B-Instruct, AASD also improves the average score.
However, it does not show an advantage in code-related tasks, possibly because the model's inherent code capabilities are relatively weak, making it unable to effectively reject the wrong draft tokens.

\begin{table*}[t]
\centering
\small
\renewcommand{\arraystretch}{1.1}
\resizebox{\textwidth}{!}{
\begin{tabular}{ccccccccc}
\hline
 & NQ & TQA & 2WikiMQA & HotpotQA & Multi-News & GovReport & RepoBench-P & LCC \\
\hline
Greedy Win Rate (\%) & 2.6 & \textbf{6.7} & 3.5 & 1.3 & 15.3 & \textbf{19.2} & 0.7 & 2.1 \\
Tie Rate (\%) & 94.4 & 87.2 & 85.6 & 89.5 & 66.6 & 72.3 & 86.4 & 82.3 \\
AASD Win Rate (\%) & \textbf{3.0} & 6.0 & \textbf{10.9} & \textbf{9.2} & \textbf{18.1} & 8.5 & \textbf{12.9} & \textbf{15.6} \\
\hline
\end{tabular}
}
\caption{\label{tab:eval} Neural evaluation results for AASD with DeepSeek-V3.}
\end{table*}
To further validate the effectiveness of AASD, we employ DeepSeek-V3 \citep{deepseekv3} as a judge model to evaluate the relative quality of responses generated by AASD and autoregressive methods.
The experimental results are shown in the table \ref{tab:eval}.
AASD demonstrates superior performance to greedy sampling across most datasets in neural evaluation (except for TQA and GovReport), especially for the code tasks.

\subsection{Improvement on Efficiency}

We evaluate generation efficiency of AASD on TQA, GovReport, and LCC datasets (one dataset for each task).
Table \ref{tab:speed} displays the results with different retrieval-based speculative decoding approaches.
Since the inputs of these tasks are long and have large length variance, input length has a great impact on the prefilling time (computation time of the first step).
To avoid the influence of this method-irrelevant factor, we only report the TPS of incremental inference (without the first step of generation for each sample).
REST adopts an independent database as the drafting source, thus having a low acceptance rate for out-of-domain tasks.
AASD outperforms other approaches both on the perspective of MAL and throughput with LLaMA3.1-8B-Instruct.
It achieves a speed-up ratio up to 2.23 compared with auto-regressive decoding.
For Qwen2.5-32B-Instruct, AASD has a significant improvement in the generation efficiency of summary and code tasks. 
However, for 2WikiMQA, as the mean output length is very short (about 8 tokens), its performance is similar to PLD.
The total time overhead for retrieving the draft candidates for each step is between 0.02 milliseconds and 2 milliseconds, which is less than 5\% of the time cost of a single step for the 8B model.
For larger target models, this overhead is negligible.

\begin{table*}[h!]
\centering
\small
\renewcommand{\arraystretch}{1.1}
\resizebox{\textwidth}{!}{
\begin{tabular}{lccccccccc}
\toprule
\multirow{2}{*}{\textbf{Method}}      & \textbf{NQ} & \textbf{TQA} & \textbf{2WikiMQA} & \textbf{HotpotQA} &  \textbf{Multi-News} &\textbf{GovReport} & \textbf{RepoBench-P} & \textbf{LCC} & \multirow{2}{*}{\textbf{Avg.}} \\
\cmidrule{2-9}
 & F1 & F1 & F1 & F1 &  ROUGE-L & ROUGE-L & Edit Sim & Edit Sim & \\
\midrule
\textbf{AASD}       & \textbf{34.54}  &  91.56   &   \textbf{43.10}   &   50.38   &    15.05      &      \textbf{19.34}       &  \textbf{58.84} & \textbf{71.03} & \textbf{47.98} \\
w/ threshold verification       & 32.05  &  \textbf{91.82}   & 41.19    &   \textbf{51.22}   &    \textbf{15.16}    &      18.69     & 56.51  & 67.75 & 46.80 \\
w/ top-\textit{k} verification     & 31.21  &   82.71  &  22.12   &   27.34   &      14.93     &      18.15       &  43.17 & 45.62 & 35.66 \\

\bottomrule
\end{tabular}
}
\caption{\label{tab:abl_res} Results for ablation study with LLaMA3.1-8B-Instruct. The best results are in bold.}
\end{table*}

\begin{table*}[h!]
\centering
\small
\renewcommand{\arraystretch}{1.1}
\resizebox{\textwidth}{!}{
\begin{tabular}{lccccccccc} 
\toprule
\multirow{2}{*}{\textbf{Method}}                   & \multicolumn{3}{c}{\textbf{2WikiMQA}} & \multicolumn{3}{c}{\textbf{GovReport}} & \multicolumn{3}{c}{\textbf{LCC}}  \\ 
\cmidrule{2-10}
     & MAL & TPS  & Speed-up           & MAL & TPS  & Speed-up                   & MAL & TPS  & Speed-up              \\
\midrule
\textbf{AASD} &  \textbf{2.37}   &  \textbf{37.18}  &  \textbf{1.85}     &  \textbf{1.97}  &  \textbf{34.00} & \textbf{1.69} &      \textbf{2.39}     & \textbf{46.20}   &    \textbf{2.23}  \\
w/o alignment sampling      &  1.80  & 36.12 &  1.80  &  1.87  & 32.93  & 1.63 &         2.25   &   42.71  &  2.07    \\
w/o conditional verification       &  2.36  &  35.86  & 1.78    &  1.86  & 32.60  & 1.62 &     2.34       &  39.02   &  1.89    \\
\bottomrule
\end{tabular}
}
\caption{\label{tab:abl_spe} Inference efficiency of AASD without alignment sampling and conditional verification.}
\end{table*}

In addition, Figure \ref{fig:radar} displays the evaluation results on SpecBench.
We report the speed-up ratio compared with autoregressive decoding.
AASD outperforms the baselines in most tasks, except for the RAG task, which further proves the effectiveness in improving decoding efficiency.

\subsection{Ablation Study}
\label{sec:ablation}

We conduct the ablation study with LLaMA3.1-8B-Instruct in this section.
We evaluate the performance of AASD with top-\textit{k} verification and fixed threshold verification.
For top-\textit{k} condition, \textit{k} is set to 5.
For threshold verification, threshold is set to 0.1.
Table \ref{tab:abl_res} shows the results.
AASD achieves the highest average score.
Using the top-k condition decreases the scores for most datasets.
For evaluation the generation efficiency, we consider AASD without alignment sampling and AASD without conditional verification.
Table \ref{tab:abl_spe} displays the results.
Both alignment sampling and conditional verification contribute to the inference acceleration.

\subsection{Comparison with Large and Small models Collaboration}
Another training-free approach is to directly use the same series of small models as the draft model \citep{leviathan2023fast,kim2023speculative}.
We compare AASD with this approach in this section.
We test the inference efficiency on NQ dataset with LLaMA2 model series \citep{touvron2023llama2openfoundation}.
Table \ref{tab:bs} displays the result.
AASD outperforms SD on generation acceleration for the 7B and 70B model.
For SD, the size of the draft model has a significant impact on inference efficiency.
A draft model that is too small will result in a low acceptance rate, while a draft model that is too large will lead to excessive draft overhead.
For a 70B target model, the 7B draft model has high quality, but its draft overhead accounts for about 80\% of the total inference overhead. Therefore, it does not significantly improve throughput.
It is difficult to obtain a draft model of appropriate size that can well balance acceptance rate and overhead.
In contrast, AASD does not have this concern.
It costs little overhead and performs well as the model size scales up.
\begin{table}[t]
\centering
\small
\renewcommand{\arraystretch}{1.1}
\resizebox{\columnwidth}{!}{
\begin{tabular}{lccccc}
\toprule
\textbf{Method}   & \textbf{$M$}   & \textbf{$M_{d}$}   & \textbf{MAL} &\textbf{TPS} & \textbf{Speed-up} \\
\midrule
\textbf{ADR}  & 7B &  -  & 1.00 & 34.32 &  1.00 \\
\textbf{SD}  & 7B &  68M  & 1.16 & 24.60 &  0.71 \\
\textbf{AASD}    & 7B & -  & \textbf{1.83} &  \textbf{50.26} & \textbf{1.46} \\
\midrule
\textbf{ADR}  & 70B &  -  &  1.00 &  5.17 & 1.00\\
\textbf{SD}  & 70B &  1B  & 1.14 & 2.68 & 0.52 \\
\textbf{SD}  & 70B &  7B  & \textbf{4.13} & 6.61 &  1.28 \\
\textbf{AASD}     & 70B &  -  & 1.83 & \textbf{7.94} & \textbf{1.53}  \\
\bottomrule
\end{tabular}
}
\caption{\label{tab:bs} Comparison with speculative decoding with big and small model collaboration. SD represents standard speculative decoding. Column $M$ and $M_{d}$ indicates the target model size and draft model size.}
\end{table}

\subsection{Case Study}
\begin{table*}[h]
\setlength{\belowcaptionskip}{0.3cm}
\centering
\small
\renewcommand{\arraystretch}{1.1}
\resizebox{\textwidth}{!}{
\begin{tabular}{p{0.18\textwidth}p{0.82\textwidth}}
\toprule
\textbf{Method} & \textbf{Label and Predictions}\\
\midrule
Ground Truth & \textbf{Mahesh Bhatt} \\
\hline
Greedy & Pooja Bhatt (\XSolidBrush)\\
AASD ($\delta$=1e-3) & Pooja Bhatt's father is \textbf{Mahesh Bhatt}. (\Checkmark) \\
AASD ($\delta$=1e-5) & Pooja Bhatt's television film Daddy was directed by her father \textbf{Mahesh Bhatt}. (\Checkmark) \\
AASD ($\delta$=1e-7) & Pooja Bhatt's television film Daddy was directed by Pooja Bhatt, starring! the director's father, played by actor Anupam Kher (\XSolidBrush)\\
\bottomrule
\end{tabular}
}
\caption{\label{tab:threshold} Comparison of generated results with AASD under different thresholds. The question is ``Who is the father of the director of film Kajraare?''}
\vspace{-10pt}
\end{table*}

Section \ref{sec:acc} shows that AASD has the potential to improve generation accuracy with a near relatively strict threshold.
In this section, we explore how the verification threshold affects the generation results through a case study.
Table \ref{tab:threshold} displays the predictions under different thresholds for a sample in 2WikiMQA.
Vanilla output fails to give the correct answer.
However, the prediction of AASD ($\delta$=1e-3 and $\delta$=1e-5) contains the ground truth.
With an appropriate threshold, AASD improves the accuracy of generation through context-aware drafting from the relevant context.
When the threshold is set to 1e-5, the logic of generated content begins to lose.
The prediction is completely messy when the threshold comes to 1e-7.
Therefore, the threshold should be large enough to keep the model output logical and fluent.
Besides, too relaxed verification may affect the model's ability to defend against attack inputs.
If the input context contain harmful information, using AASD may lead the model to incorporate it into its responses.
Therefore, in practical applications, AASD requires additional safety alignment to ensure the security of the language model.
See Appendix \ref{app:threshold} for a more comprehensive study for the impact of verification strictness.

\section{Related Work}
\textbf{Context Utilization} Using context to improve generation performance is typical in retrieval-augmented generation (RAG) scenarios.
RAG methods \citep{zheng2023take,daipromptagator,gao2023retrieval,fan2024survey,zhao2024retrieval} retrieves relevant documents based on the given input for model reference, thus improving the quality of generation. 
The naive approach \citep{ma2023query} applies the search engine as the retriever and directly combines the retrieved documents with the user query as the input for frozen LLMs. 
Most RAG methods \citep{yoran2023making,luo2023sail,asai2023self,melz2023enhancing,yan2024corrective} leveraging context on the input side to enhance generation.
Self-RAG \citep{asai2023self} introduces generating reflection tokens to enable customizing models’ behaviors for different tasks.
Speculative RAG \citep{wang2024speculative} adopts instruct-tuned draft models to drafting according to different retrieved documents and uses the target model to pick out the best draft as the final response.
Apart from them, CoG \citep{lan2023copy} proposes a encoder-based model architecture to seek suitable text spans from the context during generation.
\citet{cao2024retrieval} improves CoG through linguistic heuristics initialization and iterative self-reinforcement.

\textbf{Speculative decoding} \citet{stern2018blockwise,leviathan2023fast,xia2023speculative,xia-etal-2024-unlocking} adopt a ``drafting-verification'' pattern to lossless accelerates autoregressive decoding.
At each decoding step, a small model is used to draft the following few tokens.
Then, the target model verifies the draft in parallel and accepts tokens that are consistent with the original output.
In this way, multiple tokens can be generated in a single step.
Some works \citep{leviathan2023fast,cai2024medusa,li2024eagle} employ independent small models or additional trained modules as the draft models, while others \citep{saxena2023prompt,fu2024break,he-etal-2024-rest} retrieve drafts from a draft pool.
The draft structure has evolved from n-grams \citep{leviathan2023fast,fu2024break} to draft trees \citep{li2024eagle,wang2024opt}.
Among them, PLD \citep{saxena2023prompt} and LLMA \citep{yang2023inference} retrieve n-grams from the prompt to construct the draft.
REST \citep{he-etal-2024-rest} uses a common public data source to build a draft pool and retrieves the tree structure draft according to the last several tokens of the current sequence at each decoding step.
These retrieval-based methods avoid the hassle of additional training, but their acceleration performance is relatively lower.

\section{Conclusion}
In this paper, we proposed AASD, an alignment-augmented speculative decoding algorithm with alignment sampling and conditional verification.
Both techniques improve the input-model alignment in prompt-based speculative decoding, thus increasing the mean acceptance length and improving decoding efficiency.
Conditional verification based on a heuristic probability threshold shows the potential to enhance the generation accuracy at the token level.
AASD does not introduce additional parameters or training, making it conveniently applicable to pre-trained LMs.
It outperforms the baseline with two different models on the average score of 8 datasets in terms of generation performance and achieves a speed-up ratio of up to 2.23.

\section*{Limitations}
Due to resource constraints, we only consider applying alignment sampling on prompt-based speculative decoding in this paper.
However, alignment sampling is applicable to most existing retrieval-based speculative decoding (including REST) and most generated scenarios. Taking REST as an example, if we input the draft pool into the model in advance to obtain the output distribution of top-k tokens (or directly use the content of the model's historical output to build the draft pool), then when using REST, we can use alignment sampling to improve the alignment of the draft pool with the target model, thereby improving the draft token acceptance rate and algorithm performance.
We leave this for future work.

If the input context contains harmful information, using AASD may lead the model to incorporate it into its responses.
Therefore, in practical applications, AASD requires additional safety alignment to ensure the security of the language model.

\section*{Acknowledgments}
We want to thank all the anonymous reviewers for their valuable comments. This work was supported by the National Science Foundation of China (NSFC No. 62206194), the Natural Science Foundation of Jiangsu Province, China (Grant No. BK20220488), and the Young Elite Scientists Sponsorship Program by CAST (2023QNRC001). We also acknowledge MetaStone Tech. Co. for providing us with the software, optimization on high performance computing and computational resources required by this work.
\bibliography{custom}

\clearpage

\appendix

\section{Trade-off between Accuracy and Efficiency}
\label{app:threshold}
\vspace{-10pt}
\begin{figure}[h!]
    \centering
    \includegraphics[width=\columnwidth]{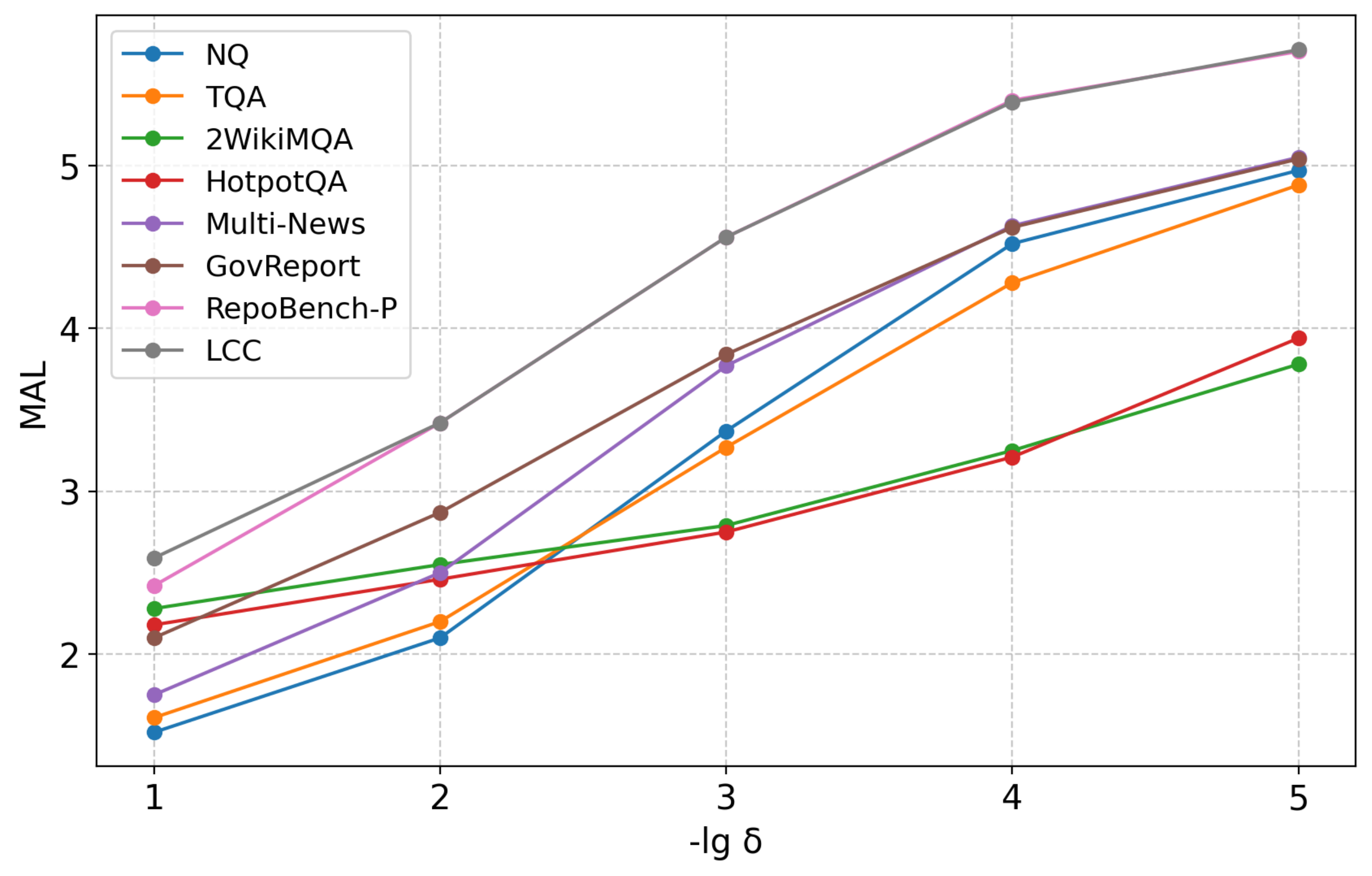}
    \caption{\label{fig:mal} Mean acceptance length of different tasks under threshold verification.}
    \vspace{-10pt}
\end{figure}
For non-strict verification, more relaxed verification conditions lead to faster reasoning, but may also cause performance degradation.
To study the trade-off between accuracy and efficiency, we conduct an experiment on LLaMA3.1-8B-Instruct \citep{llama31} with threshold verification.

\begin{figure*}[t!]
    \centering
    \includegraphics[width=\textwidth]{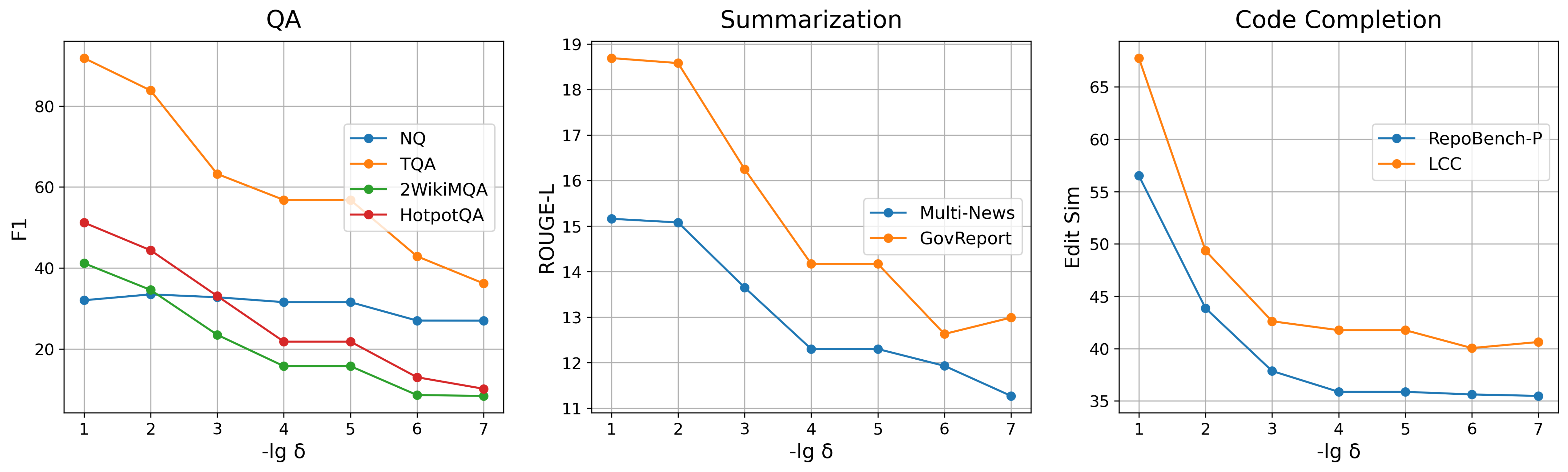}
    \caption{\label{fig:threshold} Performance of different tasks under threshold verification. The dotted lines in each figure are the baselines of the dataset with the corresponding colors.}
\end{figure*}

Figure \ref{fig:threshold} demonstrates the performance on each dataset as the threshold changes.
Note that alignment sampling is not applied.
F1 score increases with decreasing threshold for NQ while decreases for other 3 QA datasets.
The ground truth is implicit in the context for each sample in NQ.
A more relaxed verification condition makes the model tend to use the original statement in the context, which may be more accurate than the model's output in some cases, thus leading to better results.
Therefore, for high-quality context, the threshold can be lowered to increase confidence in the context for better performance and efficiency.
For both summary datasets, the overall trend of the scores decreases as the threshold decreases in the search range.
The performance drops rapidly for more complex tasks such as code completion.

Figure \ref{fig:mal} displays the mean acceptance length under different thresholds.
For all datasets, a lower threshold leading to a higher acceptance rate, thus improving the inference efficiency.

\section{Target Length and Source-Target Overlapping Rate of Each Datasets}

To explore the effectiveness of AASD in different generation scenarios, we report the target length and overlapping rate between the input and output context of each dataset used in the main experiment.
We adopts longest substring match ratio (R) to evaluate the source-target overlapping rate, which is calculated as:
\begin{equation}
     R=\frac{LCS(X,Y)}{|Y|},
\end{equation}
where $LCS$ is length of the longest common subsequence of input context $X$ and target context $Y$.
The average source-target overlapping rate and the average target length of each dataset are shown in Table \ref{tab:ol}.
Even for QA tasks (e.g., TQA, 2WikiQA and HotpotQA) where the average target length is very short, AASD demonstrates effectiveness according to the results in Table \ref{tab:main}.
On the other hand, although AASD exhibits greater advantages in tasks with high overlap, it remains effective even in low-overlap scenarios, highlighting its versatility across diverse generation tasks.

\section{The Overall Description of AASD}
\label{sec:alg}
\renewcommand{\algorithmicrequire}{ \textbf{Input:}}
\renewcommand{\algorithmicensure}{ \textbf{Output:}}
\newcommand{\RightComment}[1]{\hfill\(\triangleright\) #1}
\begin{algorithm}[h]
\caption{AASD}
\label{alg:AASD}
\begin{algorithmic}[1]
\REQUIRE ~~
Prompt $q$, LM $M$, $\alpha,\beta$, length of n-grams $n$, Maximum length of the key $l$.
\ENSURE ~~
Prediction $A_{q}$

\STATE $pool \leftarrow \{\}$ 
\FOR{$k=1$ to $l$}
\FOR{$i=0$ to $len(q)-k$}
\STATE $key \leftarrow q\left [i:i+k\right ]$
\STATE $pool\left [key\right ].append(i+k)$
\ENDFOR
\ENDFOR \RightComment{Initialize the draft pool.}
\WHILE{$ \left<eos\right>$ not in $q$}
    \STATE $\mathbb{T} \leftarrow \{q[-1]\}$ 
    \FOR{$k=l$ to $1$}
    \IF{$q\left [-k:\right ] \in pool$}
    \STATE $\mathbb{T} \leftarrow \mathbb{T} + Sample(q\left [pool\left [q\left [-k:\right ]\right ]\right ])$
    \RightComment{Conduct alignment sampling.}
    \STATE $break$
    \ENDIF 
    \ENDFOR
    \STATE $P \leftarrow M(\mathbb{T})$ 
    \STATE $y \leftarrow Verify(P,\mathbb{T},\alpha,\beta)$
    \STATE $q\leftarrow q+y$
    \FOR{$k=1$ to $l$}
    \FOR{$i=-k-len(y)$ to $-k$}
    \STATE $key \leftarrow q\left [i:i+k\right ]$
    \STATE $pool\left [key\right ].append(i+k)$
    \ENDFOR
    \ENDFOR \RightComment{Update the draft pool.}
\ENDWHILE
\STATE $A_{q} \leftarrow q$

\end{algorithmic}
\end{algorithm}

\begin{table*}[t]
\centering
\small
\renewcommand{\arraystretch}{1.1}
\resizebox{\textwidth}{!}{
\begin{tabular}{lcccccccc}
\toprule
              & \textbf{TQA}  & \textbf{2WikiQA} & \textbf{HotpotQA} & \textbf{NQ}   & \textbf{MultiNews} & \textbf{GovReport} & \textbf{LCC}   & \textbf{Repobench-P} \\
\midrule
Input length  & 9681 & 8849    & 9458     & 3612 & 7882      & 8161      & 13516 & 15300        \\
Target length & 5    & 5       & 5        & 145  & 335       & 702       & 13    & 14           \\
$R$             & 0.13 & 0.77    & 0.82     & 1.00 & 0.43      & 0.54      & 0.71  & 0.42     \\   
\bottomrule
\end{tabular}
}
\caption{\label{tab:ol} The target length and overlapping rate ($R$) between the input and output context of each dataset used in the main experiment.}
\end{table*}

\section{Efficiency under Different Batch Sizes}

Maintaining the acceleration at large batch sizes is a fundamental challenge for speculative decoding approaches.
We show speed-up ratio of AASD on SpecBench under different batch sizes in table \ref{tab:batch}.
\begin{table}[h]
\centering
\small
\renewcommand{\arraystretch}{1.1}
\resizebox{\columnwidth}{!}{
\begin{tabular}{lcccc}
\toprule
\textbf{Batch size}   & 1   & 2 & 4 & 8   \\
\midrule
\textbf{Speed-up} &1.83	& 1.65	& 1.60 &	0.96 \\
\bottomrule
\end{tabular}
}
\caption{\label{tab:batch} Speed-up ratio of AASD on Specbench under different batch sizes.}
\end{table}

Increasing the batch size does lead to a decrease in the speedup ratio for speculative decoding. 
AASD can accelerate inference with a batch size of up to 4 in our experiment environment. 

\section{Case Analysis}
\begin{figure}[h]
    \centering
    \includegraphics[width=\columnwidth]{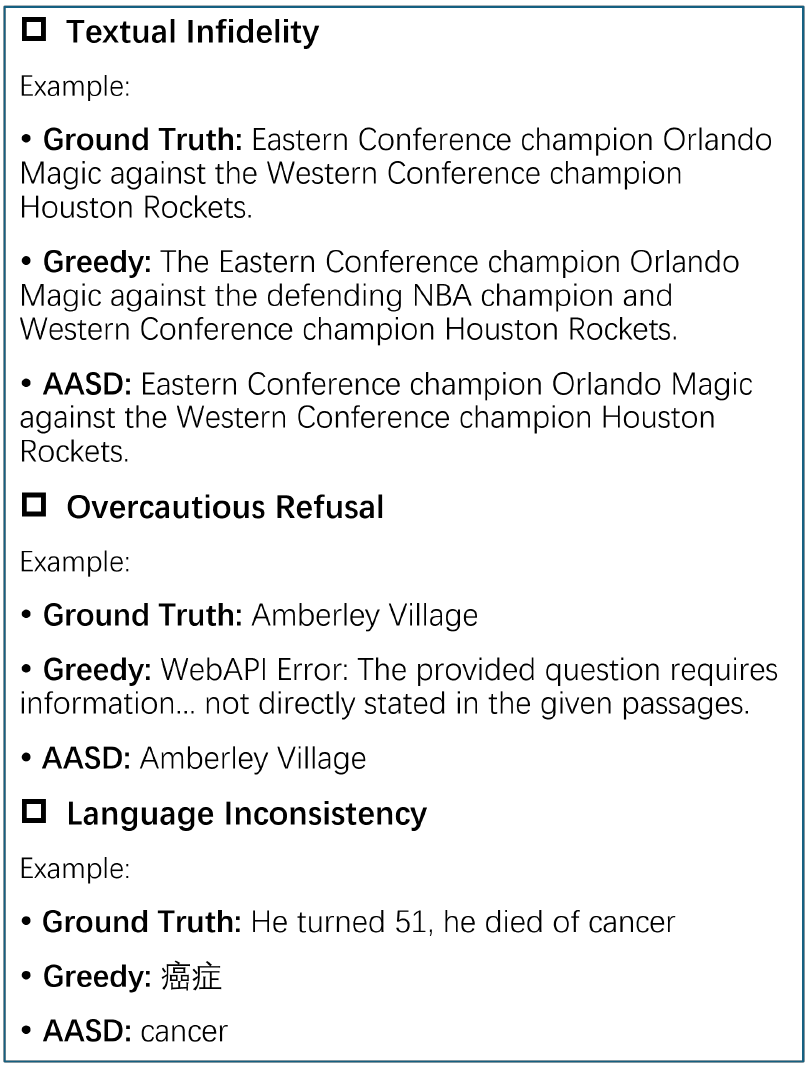}
    \caption{\label{fig:bg} Analysis for cases where AASD responses were correct while greedy sampling produced errors.}
\end{figure}

To study under what circumstances AASD can improve accuracy, we examined multiple cases where AASD responses were correct while greedy sampling produced errors, identifying three distinct failure patterns, which are shown in Figure \ref{fig:bg}

\end{document}